\documentclass[10pt,journal,compsoc]{IEEEtran}
\ifCLASSOPTIONcompsoc
  \usepackage[nocompress]{cite}
\else
  \usepackage{cite}
\fi
\ifCLASSINFOpdf
   \usepackage[pdftex]{graphicx}
   \DeclareGraphicsExtensions{.pdf,.jpeg,.png}
\else
   \usepackage[dvips]{graphicx}
   \DeclareGraphicsExtensions{.eps}
\fi

\usepackage{amssymb}
\usepackage{amsmath}
\usepackage{algorithm}
\usepackage{algorithmic}
\usepackage{url}

\newcommand{\mbf}[1]{\boldsymbol{#1}}
\newtheorem{theorem}{Theorem}

\begin{document}
%
\title{Inference in topic models: sparsity and trade-off}
%
%
%
%

\author{Khoat Than,~\IEEEmembership{Member,~IEEE,}
        Tu Bao Ho,~\IEEEmembership{Member,~IEEE} 
\IEEEcompsocitemizethanks{\IEEEcompsocthanksitem Khoat Than is with Hanoi University of Science and Technology, 1, Dai Co Viet road, Hanoi, Vietnam. Tu Bao Ho is with Japan Advanced Institute of Science and Technology, 1-1 Asahidai, Nomi, Ishikawa 923-1292, Japan.\protect\\
E-mail: khoattq@soict.hust.edu.vn}
}
\IEEEtitleabstractindextext{%
\begin{abstract}
Topic models are popular for modeling discrete data (e.g., texts, images, videos, links), and provide an efficient way to discover hidden structures/semantics in massive data. One of the core problems in this field is the posterior inference for individual data instances. This problem is particularly important in streaming environments, but is often intractable. In this paper, we investigate the use of the Frank-Wolfe algorithm (FW) for recovering sparse solutions to posterior inference. From detailed elucidation of both theoretical and practical aspects, FW exhibits many interesting properties which are beneficial to topic modeling. We then employ FW to  design fast  methods, including ML-FW, for learning latent Dirichlet allocation (LDA) at large scales.  Extensive experiments show that to reach  the same predictiveness level, ML-FW can perform tens to thousand times faster than existing state-of-the-art methods for learning LDA from massive/streaming data.  
\end{abstract}

\begin{IEEEkeywords}
Sparse topic modeling, fast inference, large-scale learning, Frank-Wolfe.
\end{IEEEkeywords}}

\maketitle

\IEEEdisplaynontitleabstractindextext

%
\IEEEpeerreviewmaketitle

\IEEEraisesectionheading{\section{Introduction}\label{sec:introduction}}

Topic modeling has been increasingly maturing to be an attractive  area. Originally motivated from textual applications, it has been going beyond far from text to touch upon many amazing applications in computer vision, bioinformatics, software engineering, forensics, to name a few. Recent development \cite{SmolaS2010,NewmanASW2009,AsuncionSW2011,MimnoHB12,Hoffman2013SVI,Broderick2013streaming} in this area enables us to easily work with big text collections or stream data.

Posterior inference is an integral part of  probabilistic topic models, e.g., latent Dirichlet allocation (LDA) \cite{BNJ03}.  It often refers to the problem of estimating the posterior distribution of latent variables, such as $\mbf{z}$ (topic indices) or $\mbf{\theta}$ \emph{(topic proportion)}, for an individual document $\mbf{d}$. Knowing $\mbf{z}$  or $\mbf{\theta}$ (or their distributions) are vital in many tasks, such as understanding individual texts, dimensionality reduction, and prediction. More importantly, posterior inference is the core step when designing efficient algorithms for learning topic models from large-scale data. Unfortunately, the problem is often intractable \cite{SontagR11}.

\subsection{The topic and  contributions in this paper}
We consider the MAP inference problem: 
\[\mbf{\theta}^* = \arg \max \Pr(\mbf{\theta, d} | \mathcal{M}),\] 
given a document $\mbf{d}$ and a model $\mathcal{M}$.
We investigate the benefits of the Frank-Wolfe algorithm (FW) by \cite{Clarkson2010} when used to do posterior inference in topic models. On one hand, this algorithm has a fast rate of convergence to optimal solutions. On the other hand, FW  can swiftly recover sparse $\mbf{\theta}$'s and  provides a way to directly trade off sparsity of solutions against quality. Those properties are essential in order to resolve large-scale settings. Note that sparsity in topic models has been receiving considerable attentions recently. FW provides a very simple way to deal with sparsity. Therefore, FW seems to have many more attractive properties than traditional inference methods. More detailed comparison is summarized in Table \ref{table 1: theoretical comparison}.

Our second contribution is the introduction of 3 novel algorithms for learning LDA at large scales: \emph{Online-FW} which borrows ideas from online learning \cite{Hoffman2013SVI}; \emph{Streaming-FW} which borrows ideas from stream learning \cite{Broderick2013streaming}; and \emph{ML-FW} which is regularized online learning. Those algorithms employ FW as the core step to do inference for individual documents, and learn LDA in a stochastic way. While Online-FW can only work with big datasets, Streaming-FW and ML-FW can work with both big collections and data streams. Extensive experiments demonstrate that those methods are much more efficient than the state-of-the-art learning methods, but keep comparable generalizability and quality. In particular, to reach  the same level of predictiveness, ML-FW can perform tens to thousand times faster than existing methods. Therefore, our study results in efficient tools for learning LDA at large scales.

\subsection{Related work}\label{sec: related work}

Various methods for  inference have been proposed such as variational Bayes  (VB) \cite{BNJ03}, collapsed variational Bayes (CVB) \cite{TehNW2007collapsed,Asuncion+2009smoothing}, collapsed Gibbs sampling (CGS) \cite{MimnoHB12,GriffithsS2004}. Sampling-based methods may converge to the underlying distributions. VB and CVB are much faster, and CVB0 \cite{Asuncion+2009smoothing} often performs  best. Although these inference methods are significant developments for topic models, they remain two common limitations that should be further studied in both theory and practice. First, there has been no theoretical  bound on convergence rate and inference quality. Second, the inferred topic proportions of documents are  dense, which requires considerable memory for storage.
\footnote{Some attempts have been initiated to speed up inference time and to attack the sparsity problem for Gibbs sampling \cite{MimnoHB12}. Sparsity in those methods does not lie in the topic proportions of documents, but lies in sufficient statistics of Gibbs samples.}

Previous researches that have attacked the sparsity problem can be categorized into two main directions. The first direction is probabilistic \cite{WilliamsonWHB2010} for which some probability distributions or stochastic processes are employed to control sparsity. The other direction is non-probabilistic for which regularization techniques are employed to induce sparsity \cite{ZhuX2011,ShashankaRS2007,LarssonU11}. Although those approaches have gained important successes, they  suffer from some severe drawbacks. Indeed, the probabilistic approach often requires extension of core topic models to be more complex, thus complicating learning and inference. Meanwhile, the non-probabilistic one often changes the objective functions of inference to be non-smooth which complicates doing inference, and requires some more auxiliary parameters associated with regularization terms. Such parameters necessarily require us to do model selection to find an acceptable setting for a given dataset, which is sometimes expensive. Furthermore, a common limitation of these two approaches is that the sparsity level of the latent representations is a priori unpredictable, and cannot be directly controlled.

There is inherently a tension between sparsity and time in the previous inference approaches. Some approaches focusing on speeding up inference \cite{BNJ03,TehNW2007collapsed,Asuncion+2009smoothing} often ignore the sparsity problem. The main reason may be that a zero contribution of a topic to a document is implicitly prohibited in some models, in which Dirichlet distributions \cite{BNJ03} or logistic function \cite{BleiL07} are employed to model latent representations of documents. Meanwhile, the approaches dealing with the sparsity problem often require more time-consuming inference, e.g., \cite{WilliamsonWHB2010,LarssonU11}.\footnote{The method by Zhu and Xing \cite{ZhuX2011} is an exception, for which inference is potentially fast. Nonetheless, their inference method cannot be applied to probabilistic topic models, since unnormalization of latent representations is required.} Note that in many practical applications, e.g., information retrieval and computer vision, fast inference of sparse latent representations of documents is of substantial significance. Hence resolving this tension is necessary.

\subsection{Roadmap} We review briefly in Section \ref{sec:post-inference} some of the most common methods for doing inference in topic models. Section \ref{sec:FW} presents the Frank-Wolfe algorithm, discusses how to employ it to topic models, and then some interesting benefits of FW. We present 3 new stochastic algorithms for learning LDA in Section \ref{sec:stochasticLDA}, and then followed by empirical evaluations in Section \ref{sec:evaluation}. Some conclusions are in the final section.

\noindent
\textsc{Notation:}

\noindent
\begin{tabular}{rl}
   $\mathcal{V}$: & a vocabulary of $V$ terms, often written as $\{1, 2,...,V\}$ \\
   $\boldsymbol{d}$: & a document represented as a count vector, \\
   & $\mbf{d} = (d_1, ..., d_V)$, where $d_j$ is the frequency of term $j$ \\
   $n_d$: & the number of different terms in $\mbf{d}$ \\
   $\ell_d$: & the length of $\mbf{d}$ \\
  $\mathcal{C}$: & a corpus consisting of $M$ documents, $ \{\boldsymbol{d}_1, ..., \boldsymbol{d}_M\}$ \\
  $\boldsymbol{\beta}_k$: & a topic which is a distribution over the vocabulary $\mathcal{V}$. \\
   & $\boldsymbol{\beta}_k = (\beta_{k1},...,\beta_{kV})^t,$ $\beta_{kj} \ge 0, \sum_{j=1}^{V} \beta_{kj} =1$ 
\end{tabular}

\noindent
\begin{tabular}{rl}
  $N_{kj}$: & the expected \# of times that term $j$ appears in topic $k$. \\
  $\lambda_{kj}$: & the variational parameter showing the contribution \\
  & of term $j$ to topic $k$. \\
  $\phi_{jk}$: & the variational parameter showing the probability \\
  & that term $j$ is generated from topic $k$. \\
  $\phi_{ik}$: & the variational parameter showing the probability \\
  & that token $i$ is generated from topic $k$. \\
  $\gamma_k$: & the variational parameter showing the expected \\
  & contribution of topic $k$. \\
  $\psi(\cdot)$: & the digamma function. \\
   $K$: & number of topics. \\	
  $\mbf{e}_i$: & the $i$th unit vector in $\mathbb{R}^K$. \\
  $\Delta_K$: & the unit simplex $\Delta_K = conv(\boldsymbol{e}_1, ..., \boldsymbol{e}_K)$ or \\ & $\Delta_K = \{\boldsymbol{x} \in \mathbb{R}^K: ||\boldsymbol{x}||_1 = 1, \boldsymbol{x} \ge 0\}$. \\
  $\mathbb{I}(x)$: & the indicator function which returns 1 if $x$ is true, \\
  & and 0 otherwise. \\
  $\nabla f$: & the gradient of function $f$.
\end{tabular}

\section{Backgrounds on posterior inference} \label{sec:post-inference}

A topic model often assumes that a  corpus is composed from $K$ topics, $\boldsymbol{\beta} = (\boldsymbol{\beta}_1, ..., \boldsymbol{\beta}_K)$. Each document $\mbf{d}$ is a mixture of those topics and is assumed to  arises from the following generative process:

For the $i^{th}$ word of $\mbf{d}$:
  \begin{itemize}
    \item[-] draw topic index $z_{i} | \mbf{\theta} \sim Multinomial(\mbf{\theta})$
    \item[-] draw word $w_{i}| z_{i}, \mbf{\beta} \sim Multinomial(\mbf{\beta}_{z_{i}})$.
  \end{itemize}
Each topic mixture $\mbf{\theta} = (\theta_{1}, ..., \theta_{K})$ represents the contributions of topics to document $\mbf{d}$, i.e., $\theta_k = \Pr(z=k | \mbf{d})$. Each $\beta_{kj} = \Pr(w=j | z=k)$ shows the contribution of term $j$ to topic $k$. Note that  $\mbf{\theta} \in \Delta_K, \mbf{\beta}_k \in \Delta_V, \forall k$. Both $\mbf{\theta}$ and $\mbf{z}$ are hidden variables and are local for each document. 

The generative process above generally describes what probabilistic latent semantic analysis (PLSA) by \cite{Hof01} is. Latent Dirichlet allocation (LDA) \cite{BNJ03} further assumes that $\mbf{\theta}$ and $\mbf{\beta}$ are samples of some Dirichlet distributions. More specifically, $\mbf{\theta} \sim Dirichlet(\alpha)$ and $\mbf{\beta}_k \sim Dirichlet(\eta)$ for any topic.

According to \cite{TehNW2007collapsed}, \emph{the  problem of posterior inference} for each document $\mbf{d}$, given a model $\{\mbf{\beta}, \alpha\}$, is to estimate the full joint distribution $p(\mbf{z}, \mbf{\theta}, \mbf{d} | \mbf{\beta}, \alpha)$. Direct estimation of this distribution is intractable, i.e., NP-hard in the worst case \cite{SontagR11} . Hence existing inference approaches use different schemes. VB, CVB, and CVB0 try to estimate the distribution by maximizing a lower bound of the likelihood $p(\mbf{d} | \mbf{\beta}, \alpha)$, whereas CGS \cite{MimnoHB12} tries to estimate $p(\mbf{z} | \mbf{d}, \mbf{\beta}, \alpha)$. We will revisit those methods briefly in the nexts subsections, with LDA as the base model.

\subsection{Variational Bayes (VB)}

VB by \cite{BNJ03} is one of the first methods to do posterior inference for LDA. The learning problem of LDA is to estimate the full joint distribution $\Pr(\mbf{z, \theta, \beta} | \mathcal{C})$ given a corpus $\mathcal{C}$. This problem is intractable in the worst case \cite{SontagR11}. To overcome intractability, VB assumes that the latent variables are independent. Specifically, we use a simpler factorized distribution $Q$ to estimate the joint distribution $\Pr(\mbf{z, \theta, \beta} | \mathcal{C})$, where

\begin{equation}
\label{eq-vb-01}
Q(\mbf{z, \theta, \beta}) = \prod_{d \in \mathcal{C}} Q(\mbf{z}_d | \mbf{\phi}_d) \prod_{d \in \mathcal{C}} Q(\mbf{\theta}_d | \mbf{\gamma}_d) \prod_k Q(\mbf{\beta}_k | \mbf{\lambda}_k).
\end{equation}
Since then, the learning problem is reduced to estimating the variational parameters $\{\mbf{\phi, \gamma, \lambda}\}$, by maximizing an evidence lower bound (ELBO) on the likelihood $\Pr(\mathcal{C} | \alpha, \eta)$, i.e.

 \begin{equation}
 \label{eq-vb-02}
 \max \mathbb{E}_{Q(\mbf{z,\theta, \beta})} \left[ \log \Pr(\mbf{z, \theta, \beta}, \mathcal{C} | \alpha, \eta) \right] + H(Q(\mbf{z, \theta, \beta})),
 \end{equation}
where $H(x)$ denotes the entropy of $x$. Note that VB implicitly assumes $\mbf{\beta}_k \sim Dir(\mbf{\lambda}_k)$.

Due to the modulo nature of VB, individual documents can be independently dealt with. Algorithm~\ref{alg:-VB} describes in details how VB estimates  $\Pr(\mbf{z, \theta} | \mbf{d}, \mbf{\beta}, \alpha)$ to do posterior inference for a document.

It is easy to observe that VB requires $O(Kn_d + K)$ to store the variational parameters for each document. Each iteration needs $O(Kn_d + K)$ arithmetic computations to update $\mbf{\gamma}$ and $\mbf{\phi}$. VB also requires computation of some expensive functions including digamma and exponent.  In particular,  for each iteration VB needs $O(Kn_d + K)$ evaluations of digamma and exponent functions. Those computations cause VB to consume significant time in practices.

\begin{table}[tp]
\centering
\begin{algorithm}[H]
    \caption{VB: variational Bayes}
      \label{alg:-VB}
   \begin{algorithmic}
      \STATE {\bfseries Input: } document $\boldsymbol{d}$, model  $\{\mbf{\lambda}, \alpha\}$.
      \STATE {\bfseries Output:} $\mbf{\phi}$.
      \STATE Initialize $\mbf{\phi}$ randomly.
      \FOR{ $\ell = 0, ..., \infty$}
            \STATE $\gamma_k :=  \alpha + \sum_{d_j > 0} \phi_{jk} d_j$
      		\STATE $\phi_{jk} \propto  \exp \psi(\gamma_{k}) .\exp[ \psi(\lambda_{kj}) - \psi(\sum_t \lambda_{kt}) ]$
      \ENDFOR
   \end{algorithmic}
\end{algorithm} 
\centering
\begin{algorithm}[H]
\caption{CVB: collapsed variational Bayes}
       \label{alg:-CVB}
    \begin{algorithmic}
       \STATE {\bfseries Input: } document $\boldsymbol{d}$, model  $\{\mbf{N}, \alpha, \eta\}$.
       \STATE {\bfseries Output:} $\mbf{\phi}$.
       \STATE Initialize $\mbf{\phi}$ randomly.
       \FOR{ $\ell = 0, ..., \infty$}
       		\FOR{the $i$th token $z_i$ in $\mbf{d}$}
       		\STATE $\gamma^{-i}_{k} := \alpha + \sum_{t \neq i} \phi_{tk}$ 
       		\STATE $V^{-i}_{k} := \sum_{t \neq i} \phi_{tk} (1-\phi_{tk})$
       		\STATE $N^{-i}_{kz_i} := N^{-i}_{kz_i} +  \phi_{ik}$ 
       		\STATE $a_k^{-i} := \sum_t N_{kt}^{-i} $
       		\STATE $X :=  - \frac{V_k^{-i}}{2(\gamma_{k}^{-i})^2} - \frac{V_{kz_i}^{-i}}{2(N_{kz_i}^{-i} + \eta)^2} + \frac{V_k^{-i}}{2(a_{k}^{-i} + V\eta)^2} $
       		\STATE $\phi_{ik} \propto \gamma_{k}^{-i} (N_{kz_i}^{-i} + \eta) (a_{k}^{-i} + V\eta)^{-1}  \exp X$
       		\ENDFOR
       \ENDFOR
    \end{algorithmic}
\end{algorithm} 
\centering
\begin{algorithm}[H]
\caption{CVB0: a fast variant of CVB}
   \label{alg:-CVB0}
\begin{algorithmic}
   \STATE {\bfseries Input: } document $\boldsymbol{d}$, model  $\{\mbf{N}, \alpha, \eta\}$.
   \STATE {\bfseries Output:} $\mbf{\phi}$.
   \STATE Initialize $\mbf{\phi}$ randomly.
   \FOR{ $\ell = 0, ..., \infty$}
   		\FOR{the $i$th token $z_i$ in $\mbf{d}$}
   		\STATE $\gamma^{-i}_{k} := \alpha + \sum_{t \neq i} \phi_{tk} $
   		\STATE $N^{-i}_{kz_i} := N^{-i}_{kz_i} +  \phi_{ik}$  
   		\STATE $a_k^{-i} := \sum_t N_{kt}^{-i} $
   		\STATE $\phi_{ik} \propto \gamma_{k}^{-i} (N_{kz_i}^{-i} + \eta) (a_{k}^{-i} + V\eta)^{-1}$
   		\ENDFOR
   \ENDFOR
\end{algorithmic}
\end{algorithm} 
\centering
\begin{algorithm}[H]
    \caption{CGS: collapsed Gibbs sampling}
       \label{alg:-CGS}
    \begin{algorithmic}
       \STATE {\bfseries Input: } document $\boldsymbol{d}$, model  $\{\mbf{\lambda}, \alpha\}$.
       \STATE {\bfseries Output:} $\mbf{\phi}$.
       \STATE Initialize $\mbf{z}$ randomly.
       \STATE Discard $B$ burn-in sweeps.
       \FOR{ $\ell = 1, ..., S$ samples}
       		\FOR{the $i$th token $z_i$ in $\mbf{d}$}
       		\STATE $\gamma^{-i}_{k} := \alpha + \sum_{t \neq i} \mathbb{I}(z_t = k) $
       		\STATE $\phi_{ik} \propto \gamma_{k}^{-i}  \exp [ \psi(\lambda_{kz_i}) - \psi(\sum_t \lambda_{kt}) ]$
       		\STATE Sample $z_i$ from $Multinomial(\mbf{\phi}_i)$
       		\ENDFOR
       \ENDFOR
    \end{algorithmic}
\end{algorithm} 


\begin{algorithm}[H]
   \caption{FW: Frank-Wolfe}
   \label{alg:-Frank-Wolfe}
\begin{algorithmic}
   \STATE {\bfseries Input: } document $\boldsymbol{d}$, model $\mbf{\beta}$, objective function  $f(\mbf{\theta}) = \sum_{j} d_j \log \sum_{k=1}^K \theta_k \beta_{kj}$.
   \STATE {\bfseries Output:} $\boldsymbol{\theta}$  that maximizes $f(\boldsymbol{\theta})$ over $\Delta_K$.
   \STATE Pick as $\boldsymbol{\theta}_{0}$ the vertex of $\Delta_K$ with largest $f$ value.
   \FOR{ $\ell = 0, ..., \infty$}
   \STATE $i' := \arg \max_i   \nabla f(\boldsymbol{\theta}_{\ell})_{i}$;
   \STATE $\alpha := 2 / (\ell+3)$;
   \STATE $\boldsymbol{\theta}_{\ell +1} := \alpha \boldsymbol{e}_{i'} +(1-\alpha)\boldsymbol{\theta}_{\ell}$.
   \ENDFOR
\end{algorithmic}
\end{algorithm}
\end{table}

\subsection{Collapsed variational Bayes (CVB)}

Instead of using a full factorized distribution, CVB by \cite{TehNW2007collapsed} uses 
\begin{equation}
 \label{eq-cvb-03}
Q(\mbf{z, \theta, \beta}) = Q(\mbf{\theta, \beta} | \mbf{z, \gamma, \lambda}) \prod_{d \in \mathcal{C}} Q(\mbf{z}_d | \mbf{\phi}_d) 
\end{equation}
to approximate $\Pr(\mbf{z, \theta, \beta} | \mathcal{C})$. The resulting problem is

\begin{equation}
 \label{eq-cvb-04}
\max \mathbb{E}_{Q(\mbf{z}) Q(\mbf{\theta, \beta} | \mbf{z})} \left[ \log \Pr(\mbf{z, \theta, \beta}, \mathcal{C} | \alpha, \eta) \right] + H(Q(\mbf{z}) Q(\mbf{\theta, \beta} | \mbf{z})).
\end{equation}

We maximize the objective function with respect to $Q(\mbf{\theta, \beta} | \mbf{z})$ first and followed by $Q(\mbf{z})$ until convergence. Note that CVB can give better approximations than VB because of maintaining the dependency between $\mbf{z}$ and $(\mbf{\theta, \beta})$. Borrowing ideas from Gibbs sampling \cite{GriffithsS2004}, CVB exploits individual tokens in documents to do inference. As an example, while VB maintains a variational distribution $\mbf{\gamma} = (\gamma_1, ..., \gamma_K)$ for each document, CVB maintains a $\mbf{\gamma}$ for each token. Such a deeper treatment probably helps CVB work better than VB.

When adapting to inference for a specific document $\mbf{d}$, we find that CVB in fact tries to estimate $\Pr(\mbf{z} | \mbf{d}, \alpha, \eta)$ which is simpler than $\Pr(\mbf{z, \theta} | \mbf{d}, \alpha, \eta)$ in VB. However, posterior inference by CVB is not local for a particular document, and requires some updates to global variables. Details of posterior inference by CVB is presented in Algorithm~\ref{alg:-CVB}. Note that $N_{kj}$ plays a similar role with $\lambda_{kj}$ in VB.

In comparison with VB, CVB requires significantly more computations and memory for storing temporary parameters. Since CVB works with individual tokens in a document, memory for the variational parameters is $O(K\ell_d)$ where $\ell_d$ denotes the number of tokens in document $\mbf{d}$. Note that we often have $\ell_d \ge n_d$. CVB further needs to maintain the variance vector ($V^{-i}$) for each token which also requires a memory of $O(K\ell_d)$. From those observations, one can realize that each iteration of CVB requires $O(K\ell_d)$ computations.

One important property of CVB is that each update for the local variables w.p.t a token requires some modifications to the global variables ($\mbf{N}$). It may help the model update more quickly as observing individual tokens. Nonetheless, this property is not ideal for some practical cases, such as parallel/distributed inference for individual documents, as communication overhead will be very high.

\subsection{Fast collapsed variational Bayes (CVB0)}

CVB0 \cite{Asuncion+2009smoothing} is an improved version of CVB. The update for $\phi_{ik}$ in CVB makes uses of a second order Taylor extension, and is quite involved. Asuncion et al. \cite{Asuncion+2009smoothing} propose to use only the zero order information for approximation to make the update of $\phi_{ik}$ significantly simpler. Algorithm~\ref{alg:-CVB0} shows details of CVB0 for doing posterior inference for a given document.

Similar with CVB, we still have to make some updates to global variables $(N_{kj})$ when doing inference for individual documents in CVB0. Nonetheless, CVB0 does not have to maintain any variance for individual tokens. This property helps CVB0 much more efficient than CVB in both computation and memory.

Due to its simplicity, CVB0 requires much less computations and storage than the original CVB. No computation of exponents or digamma function is necessary. By a careful enumeration, we find that the complexity of CVB0 in both computation and memory is  $O(K\ell_d)$. Similar with CVB, we still need to do some modifications to global variables when doing local inference for individual documents.

\subsection{Collapsed Gibbs sampling (CGS)}

Originally, CGS was proposed by \cite{GriffithsS2004} for learning LDA from data. It recently has been successfully adapted to posterior inference for individual documents by \cite{MimnoHB12}. It tries to estimate $\Pr(\mbf{z} | \mbf{d}, \alpha, \eta)$ by iteratively resampling the topic indicator at each token in $\mbf{d}$ from the conditional distribution over that position given the remaining topic indicator variables ($\mbf{z}^{-i}$):

\begin{equation}
\label{eq-cgs-05}
 \Pr(z_i = k | \mbf{z}^{-i}) \propto \left( \alpha + \sum_{t \neq i} \mathbb{I}(z_t = k) \right)   \exp [ \psi(\lambda_{kz_i}) - \psi(\sum_t \lambda_{kt}) ].
\end{equation}

Note that this adaptation makes the inference more local, i.e., posterior inference for a document does not need to modify any global variable. This property is similar with VB, but very different with CVB and CVB0. Details are presented in Algorithm~\ref{alg:-CGS}.

To take a random sample, CGS needs $O(K\ell_d)$ computations to compute all $\phi_{ik} = \Pr(z_i = k | \mbf{z}^{-i})$. Note that CGS also needs $O(K\ell_d)$ evaluations of exponent and digamma functions which are  expensive. In total, CGS requires $O((S+B)K\ell_d)$ computations for the whole sampling procedure with $B$ burn-in sweeps and $S$ samples. Storing $\mbf{\phi}$ requires $O(K\ell_d)$ memory. 

\section{The Frank-Wolfe algorithm for posterior inference} \label{sec:FW}
This section reviews the Frank-Wolfe algorithm for concave maximization over simplex. We then discuss how to employ it to do inference of theta in LDA. Its interesting properties will be discussed and compared with common inference methods. 

\subsection{Concave maximization over simplex and sparse approximation}

Consider a concave function $f(\boldsymbol{\theta}): \mathbb{R}^K \rightarrow \mathbb{R}$ which is twice differentiable over $\Delta_K$. We are interested in the following problem, \emph{concave maximization over the unit simplex},
\begin{equation}\label{eq-fw-01}
    \boldsymbol{\theta}^* = \arg \max_{\boldsymbol{\theta} \in \Delta_K} f(\boldsymbol{\theta})
\end{equation}

Convex/concave optimization has been extensively studied in the optimization literature. There has been various excellent results such as \cite{Nesterov2005,Lan2012}. However, we are interested in sparse approximation algorithms specialized for problem (\ref{eq-fw-01}). More specifically, we focus on the Frank-Wolfe algorithm \cite{Clarkson2010}.

Loosely speaking, the Frank-Wolfe algorithm is an approximation one for  problem (\ref{eq-fw-01}). Starting from a vertex of the simplex $\Delta_K$, it iteratively selects the most potential vertex of $\Delta_K$ to change the current solution closer to that vertex in order to maximize $f(\boldsymbol{\theta})$. Details are presented in Algorithm~\ref{alg:-Frank-Wolfe}. It has been shown that the algorithm converges at a linear rate to the optimal solution. Moreover, at each iteration, the algorithm finds a provably good approximate solution lying in a face of $\Delta_K$.

\begin{theorem}\cite{Clarkson2010} \label{thm-FW}
Let $f$ be a continuously differentiable, concave function over $\Delta_K$, and denote $C_f$ be the largest constant so that $\forall \boldsymbol{\theta}, \boldsymbol{\theta}' \in \Delta_K, a \in [0, 1] $ we have $f(a \boldsymbol{\theta}' + (1-a)\boldsymbol{\theta}) \ge f(\boldsymbol{\theta}) + a(\boldsymbol{\theta}' - \boldsymbol{\theta})^t \nabla f(\boldsymbol{\theta}) - a^2 C_f$. After $\ell$ iterations, the Frank-Wolfe algorithm finds a point $\boldsymbol{\theta}_{\ell}$ on an $(\ell+1)-$dimensional face of $\Delta_K$ such that
\begin{equation}
\max_{\boldsymbol{\theta} \in \Delta_K} f(\boldsymbol{\theta}) - f(\boldsymbol{\theta}_{\ell}) \le \frac{4C_f}{(\ell +3)}.
\end{equation}
\end{theorem}

It is worth noting some observations about the algorithm:
\begin{itemize}
  \item[-] It achieves a linear rate of convergence, and has provable bounds on the goodness of approximate solutions. These are crucial for practical applications.
  \item[-] Overall running time mostly depends on how complicated $f$ and $\nabla f$ are.
  \item[-] It provides an explicit bound on the dimensionality of the face of $\Delta_K$ in which an approximate solution lies. After $\ell$ iterations, Theorem \ref{thm-FW} ensures that at most $\ell+1$ out of $K$ components of $\boldsymbol{\theta}_{\ell}$ are non-zero. 
  \item[-] It is easy to directly control the sparsity level of $\mbf{\theta}$ by trading off sparsity against quality. The fewer the number of iterations, the sparser the solution. This characteristic makes the algorithm very attractive for resolving high dimensional problems.
\end{itemize}

\subsection{How to employ FW in topic models}\label{sec:Employment-FW}

Posterior inference for a document in LDA and many models often relates to the latent variables $\mbf{z}$ and $\mbf{\theta}$. We sometimes want to know the full joint distribution $\Pr(\mbf{z, \theta | d})$, or $\Pr(\mbf{z | d})$, or $\Pr(\mbf{\theta | d})$, or even individuals $\mbf{z}$ or $\mbf{\theta}$.  Estimation of individuals $\mbf{z}$ or $\mbf{\theta}$ is often maximum a posteriori (MAP).

Here we discuss how to do inference of $\mbf{\theta}$ using FW. Note that one can make approximation to $\Pr(\mbf{z | d})$ from $\mbf{\theta}$ and vice versa.

\subsubsection{MAP inference of $\mbf{\theta}$}

We now consider LDA and the MAP estimation of topic mixture for a given document $\mbf{d}$:
\begin{equation} \label{eq-map-1}
\mbf{\theta}^* = \arg \max_{\mbf{\theta} \in \Delta_K} \Pr(\mbf{\theta}, \mbf{d}|\mbf{\beta},\alpha) = \arg \max_{\mbf{\theta} \in \Delta_K} \Pr(\mbf{d}|\mbf{\theta},\mbf{\beta}) \Pr(\mbf{\theta}|\alpha).
\end{equation}

For a given document $\boldsymbol{d}$, the probability that a term $j$ appears in $\boldsymbol{d}$ can be expressed as $\Pr(w = j | \boldsymbol{d}) = \sum_{k=1}^K \Pr(w=j | z=k).\Pr(z=k | \boldsymbol{d}) = \sum_{k=1}^K  \beta_{kj} \theta_k$. Hence the log likelihood of $\boldsymbol{d}$ is
\begin{eqnarray}
\nonumber
& & \log \Pr(\boldsymbol{d} | \mbf{\theta},\mbf{\beta}) = \log \prod_{j} \Pr(w=j | \boldsymbol{d})^{d_j} \\ 
&=& \sum_{j } d_j \log \Pr(w=j | \boldsymbol{d}) = \sum_{j} d_j \log \sum_{k=1}^K \theta_k \beta_{kj}.
\end{eqnarray}
Remember that the density of the $K$-dimensional Dirichlet distribution with parameter $\alpha$ is $p(\mbf{\theta} | \alpha) \propto \prod_{k=1}^{K} \theta_k^{\alpha -1}$. Therefore problem (\ref{eq-map-1}) is equivalent to the following:
\begin{equation} \label{eq-map-2}
\mbf{\theta}^* = \arg \max_{\mbf{\theta} \in \Delta_K} \sum_j d_j \log\sum_{k = 1}^K\theta_k\beta_{kj} + (\alpha - 1)\sum_{k = 1}^K \log\theta_k.
\end{equation} 

When $\alpha=1$, it is easy to show that problem (\ref{eq-map-2}) is concave. Hence we can employ FW to efficiently solve for $\mbf{\theta}$. In other words, FW can be used to find $\mbf{\theta}^*$ by maximizing the function $\sum_j d_j \log\sum_{k = 1}^K\theta_k\beta_{kj}$ over the unit simplex. 

By using Algorithm \ref{alg:-Frank-Wolfe} to do inference, we implicitly assume  that $\mbf{\theta}^*$ follows  the distribution $Dirichlet(1)$. Another interpretation is that we remove the Dirichlet prior over $\mbf{\theta}$. This seems to be strange and uncommon. No prior endowment over $\mbf{\theta}$ might cause some overfittings in practice \cite{BNJ03}. However, we will show that such an inference way provides us many practical benefits, and that there is an \emph{implicit sparse prior} over topic mixtures to avoid overfitting as discussed in the next subsection.

\subsubsection{Recovery of $\mbf{z}$ from $\mbf{\theta}$ and vice versa}
We can easily make a connection of $\mbf{\theta}$ and $\mbf{z}$. Note that estimation of $\mbf{z}$ is intractable in the worst case \cite{SontagR11}. Instead, we discuss a connection of $\mbf{\theta}$ and the distribution of $\mbf{z}$, as it is enough for deriving various fast algorithms for learning topic models which will be discussed in Section \ref{sec:stochasticLDA}.

Denote $\phi_{jk} = \Pr(z=k | w=j, \mbf{d})$ the probability that topic $k$ generates term $j$ in document $\mbf{d}$. Then it connects to $\mbf{\theta}$ by the following formula \cite{Asuncion+2009smoothing}

\begin{equation} \label{eq-map-10}
\phi_{jk} \propto \theta_k \beta_{kj}.
\end{equation}
When  further assuming $\mbf{\beta}$ to be a random variable, we have

\begin{equation} \label{eq-map-11}
\phi_{jk} \propto \theta_k \exp \mathbb{E}_Q (\log \beta_{kj}).
\end{equation}
If both $\mbf{\beta}$ and $\mbf{\theta}$ are random variables as in LDA, we have

\begin{equation} \label{eq-map-12}
\phi_{jk} \propto \exp\mathbb{E}_Q (\log \theta_k) .\exp \mathbb{E}_Q (\log \beta_{kj}),
\end{equation}
where $Q$ is a certain distribution. Sometimes $Q$ is a variational distribution of $\Pr(\mbf{z, \theta, \beta})$, but in some other situations $Q$ is the distribution of $(\mbf{z}^{-i}, \mbf{\theta, \beta})$ for some token $i$ removed.

Note that the expectations in (\ref{eq-map-12}) are often intractable to compute, because both $\mbf{\beta}$ and $\mbf{\theta}$ are hidden. Some popular approaches to deal with these quantities base on VB \cite{BNJ03} and CGS \cite{GriffithsS2004}. The formulas of $\phi$ in Algorithms~\ref{alg:-VB}--\ref{alg:-CVB0} are the results of different approaches to approximate the intractable expectations in (\ref{eq-map-12}), and provide some specific ways to approximate the distribution of $\mbf{z}$ given $\mbf{\theta}$.

We can make an approximation to $\mbf{\theta}$ once having known $\mbf{\phi}$. Indeed, we observe that $\mbf{\gamma}$ in Algorithms~\ref{alg:-VB}--\ref{alg:-CVB0} plays a role as sufficient statistics for the Dirichlet distribution over $\mbf{\theta}$. Hence, we can use the following approximation 

\begin{equation} \label{eq-map-13}
 \theta_k = \frac{\gamma_k}  {\sum_{t=1}^K \gamma_t}
\end{equation}

\subsection{Benefits from FW}

In this section we elucidate the main benefits of using FW, accompanied by a comparison with existing methods for posterior inference. The benefits come from both theoretical and practical perspectives. Table~\ref{table 1: theoretical comparison} summarizes the main properties of the inference methods of interests. 

\begin{table*}[tp]
\caption{Theoretical comparison of 5 inference methods, given a document $\mbf{d}$ and model $\mathcal{M}$ with $K$ topics. ML denotes maximizing the likelihood, ELBO denotes maximizing an evidence lower bound on the likelihood. $L$ denotes  the number of iterations.  `-' denotes \emph{`no'} or \emph{`unspecified'}.}
\begin{center}
\begin{tabular}{llllll}
\hline
Method & FW & VB & CVB & CVB0 & CGS \\
\hline
Posterior probability & $\Pr(\mbf{\theta, d} | \mathcal{M})$ & $\Pr(\mbf{\theta, z, d} | \mathcal{M})$ & $\Pr(\mbf{z, d} | \mathcal{M})$ & $\Pr(\mbf{z, d} | \mathcal{M})$ & $\Pr(\mbf{z, d} | \mathcal{M})$ \\
Approach & ML & ELBO & ELBO & ELBO & Sampling \\
Sparse solution & Yes & - & - & - & Yes \\
Sparsity control & direct & - & - & - & -  \\
Trade-off: & & & & &  \\
\,\,\,\, sparsity vs. quality & Yes & - & - & - & - \\
\,\,\,\, sparsity vs. time & Yes & - & - & - & - \\
Quality bound & Yes & -& -& -& - \\
Convergence rate & $O(1/{L})$ & - & -& -& - \\
Iteration complexity & $O(K. n_d)$ & $O(K. n_d)$ & $O(K. \ell_d)$ & $O(K. \ell_d)$ & $O(K. \ell_d)$ \\
Storage & $O(K)$ & $O(K. n_d)$ & $O(K. \ell_d)$ & $O(K. \ell_d)$ & $O(K. \ell_d)$ \\
$Digamma$ evaluations & 0 & $O(K.n_d)$ & 0 & 0 & $O(K.n_d)$ \\
$Exp$ or $Log$ evaluations  & $O(K.n_d)$ & $O(K.n_d)$ & $O(K. \ell_d)$ & 0 & $O(K.n_d)$ \\
Modification on global variables & No & No & Yes & Yes & No \\
\hline
\end{tabular}
\end{center}
\label{table 1: theoretical comparison}
\end{table*}

\subsubsection{Complexity and quality of inference}

One can easily observe that the initialization step and selection of a maximum gradient direction step are  most expensive in Algorithm~\ref{alg:-Frank-Wolfe}. Initialization requires $K$ evaluations of $f(\mbf{\theta})$ with respect to $K$ vertices of the simplex $\Delta_K$. For  $f(\mbf{\theta}) = \sum_j d_j \log\sum_{k = 1}^K\theta_k\beta_{kj}$, we need $O(Kn_d)$ computations to do the initialization. Taking $K$ partial differentials from $f$ and then finding the maximal one also need $O(Kn_d)$. As a consequence, $O(Kn_d)$ computations are sufficient to do an iteration for FW.

FW requires a modest amount of memory for storage, which is $O(K)$ for maintaining the solution and gradient. Such a memory consumption is significantly less than VB, CVB, CVB0, and CGS as Table~\ref{table 1: theoretical comparison} demonstrates. Therefore FW is expected to be much more efficient than other  methods in both memory and computation.

Theorem~\ref{thm-FW} suggests that FW converges very fast to the optimal solution. After $\ell$ iterations, it finds an approximate solution $\mbf{\theta}_\ell$ which is provably good, with a bounded error of $4C_f / (\ell + 3)$ in inference quality. This property of FW is very different from existing methods. To the best of our knowledge, no theory has been established  to see the convergence rate and inference quality of VB, CVB, CVB0, and CGS. Hence in practices, we are not sure about the quality of posterior inference by VB, CVB, CVB0, and CGS. In these theoretical aspects, FW behaves better than existing methods.

\subsubsection{Managing sparsity level and trade-off}

Good solutions are often necessary for practical applications. In practice, we may have to spend intensive time and significant memory to search such solutions. This sometimes is not necessary or impossible in limited time/memory settings. Hence one would prefer to trading off quality of solutions against time/memory.

Searching for sparse solutions is a common approach in Machine Learning to reduce memory for storage and efficient processing. Most previous works have tried to learn sparse solutions by imposing regularization which induces sparsity, e.g., L1 regularization \cite{ZhuX2011},  \cite{WangXLC2011} and entropic regularization \cite{ShashankaRS2007}. Nevertheless, those techniques are severely limited in the sense that we cannot directly control the sparsity level of solutions (e.g., one cannot decide how many non-zero components solutions should  have). In other words, the sparsity level of solutions is  a priori unpredictable. This limitation makes regularization techniques inferior in memory limited settings. It is also the case with other works that employ some probabilistic distributions to induce sparsity  \cite{WilliamsonWHB2010,WangB2009} or that exploits sparsity of sufficient statistics of Gibbs samples \cite{MimnoHB12}.

Unlike prior approaches, FW naturally provides a principled way to control sparsity. Theorem \ref{thm-FW} implies that if stopped at the $L$th iteration, the inferred solution has at most $L+1$ non-zero components. Hence one can control sparsity level of solutions by simply limiting the number of iterations. It means that we can predict a priori how sparse and how good the inferred solutions are. Less iterations, sparser (but probably worse) solutions of inference. Besides, we can trade off sparsity against inference time. More iterations imply more necessary time and probably denser solutions.

\subsubsection{Implicit prior over $\boldsymbol{\theta}$}

Note that FW allows us to easily trade off sparsity of solutions against quality and time. If one insists on solutions with at most $t$ nonzero components, the inference algorithm can be modified accordingly. In this case, it mimics that one is trying to find a solution to the problem $\max_{\boldsymbol{\theta} \in \Delta_K} \{f(\boldsymbol{\theta}): ||\boldsymbol{\theta}||_0 \le t\}$. We remark a well-known fact that the constraint $||\boldsymbol{\theta}||_0 \le t$ is equivalent to addition of a penalty term $\lambda.||\boldsymbol{\theta}||_0$ to the objective function \cite{Murray+1981}, for some constant $\lambda$. Therefore, one is trying to solve for 
\begin{eqnarray}
\nonumber
\boldsymbol{\theta}^* &=& \arg \max_{\boldsymbol{\theta} \in \Delta_K} \{f(\boldsymbol{\theta})- \lambda.||\boldsymbol{\theta}||_0 \} = \arg \max_{\boldsymbol{\theta} \in \Delta_K} P(\boldsymbol{d}| \boldsymbol{\theta}).P(\boldsymbol{\theta}) \\
\nonumber
&=& \arg \max_{\boldsymbol{\theta} \in \Delta_K} P(\boldsymbol{\theta}| \boldsymbol{d}),
\end{eqnarray}
where $p(\boldsymbol{\theta}) \propto \exp(-\lambda.||\boldsymbol{\theta}||_0)$. Notice that the last problem, $\boldsymbol{\theta}^*= \arg \max_{\boldsymbol{\theta} \in \Delta_K} P(\boldsymbol{\theta}| \boldsymbol{d})$, is an MAP inference problem. Hence, these observations basically show that inference by Algorithm~\ref{alg:-Frank-Wolfe} for sparse solutions mimics MAP inference. As a result, there exists an implicit prior, having density function $p(\boldsymbol{\theta}; \lambda) \propto \exp(-\lambda.||\boldsymbol{\theta}||_0)$, over latent topic proportions.

\section{Stochastic algorithms for learning LDA} \label{sec:stochasticLDA}

We have seen many interesting properties of FW. In this section, we show the simplicity of using FW to design efficient algorithms for learning topic models at large scales. More specifically, we present 3 different ways to encode FW as an internal step into online learning \cite{Hoffman2013SVI} and stream learning \cite{Broderick2013streaming}. Those encodings result in 3 novel methods which are fast and effective.

\subsection{Online-FW for learning LDA from large corpora}

Hoffman et al. \cite{Hoffman2013SVI} show that LDA can be learned efficiently in a stochastic manner. Note that the batch VB by \cite{BNJ03} learns LDA by iteratively maximizing an ELBO on the data likelihood using coordinate ascent. Each iteration of the batch VB requires to access all the available training data. Such a requirement causes batch VB to be impractical for large corpora or stream environments.

Fortunately, a simple modification  can help us learn topic models in an online fashion. Indeed, \emph{stochastic variational inference} (SVI) by \cite{Hoffman2013SVI} learns LDA iteratively from a corpus of size $D$ as follows:

\begin{itemize}
	\item[-] Sample a set $\mathcal{C}_t$ consisting of $S$ documents. Use Algorithm~\ref{alg:-VB} to do posterior inference for each document $\mbf{d} \in \mathcal{C}_t$, given the global variable $\mbf{\lambda}^{(t - 1)}$ in the last step, to get variational parameters $\mbf{\phi}_d$.
	\item[-] For each $k \in \{1, 2, ..., K\}$, form an intermediate global variable $\hat{\mbf{\lambda}}_k$ for $\mathcal{C}_t$ by
		\begin{equation}
		  \label{eq-14-ovb}
		  \hat{\mbf{\lambda}}_k = \eta + \frac{D}{S} \sum_{\mbf{d} \in \mathcal{C}_t} \sum_j d_j \phi_{djk}
		\end{equation}
	\item[-] Update the global variable to be a weighted average of  $\hat{\mbf{\lambda}}$ and $\mbf{\lambda}^{(t - 1)}$ by 
			\begin{equation}
			  \label{eq-15-ovb}
			  \mbf{\lambda}^{(t)} := (1-\rho_t) \mbf{\lambda}^{(t-1)} + \rho_t \hat{\mbf{\lambda}}.
			\end{equation}
\end{itemize}

$\rho_t$ is called the step size of the learning algorithm, and should satisfy two conditions: $\sum_t^\infty \rho_t = \infty$ and $\sum_t^\infty \rho^2_t$ is finite. Those two conditions are to assure that the learning algorithm will converge to a stationary point. In practice, we often choose
\[
  \rho_t = (\tau + t)^{-\kappa}
\]
where $\kappa \in (0.5, 1]$ is the forgeting rate which determines how fast the algorithm forgets past observations, and $\tau$ is a positve constant.

It is easy to modify SVI to employ FW instead of VB. Remember that FW infers a vector $\mbf{\theta}$, but VB infers a matrix $\mbf{\phi}$. Fortunately, equation~(\ref{eq-map-11}) shows that we can recover $\mbf{\phi}$ from $\mbf{\theta}$. Therefore, we arrive at a novel algorithm (namely, Online-FW) for learning LDA stochastically as described in Algorithm~\ref{alg:-online-FW}. 

\begin{algorithm}[tp]
   \caption{\textsf{Online-FW} for learning LDA}
   \label{alg:-online-FW}
\begin{algorithmic}
   \STATE {\bfseries Input: } training data $\mathcal{C}$ with $D$ documents, hyperparameter $\eta$
   \STATE {\bfseries Output:} $\mbf{\lambda}$
   \STATE Initialize $\mbf{\lambda}^{(0)}$ randomly
   \FOR{ $t = 1, ..., \infty$}
	   \STATE Sample a set $\mathcal{C}_t$ consisting of $S$ documents. 
	   \STATE Use Algorithm~\ref{alg:-Frank-Wolfe} to do posterior inference for each document $\mbf{d} \in \mathcal{C}_t$, given the global variable $\mbf{\beta}^{(t-1)} \propto \mbf{\lambda}^{(t - 1)}$ in the last step, to get topic mixture $\mbf{\theta}_d$. Then compute $\mbf{\phi}_d$ as 
		   \begin{equation} 
		   \phi_{djk} \propto \theta_{dk}  \beta_{kj}.
		   \end{equation}
	   \STATE For each $k \in \{1, 2, ..., K\}$, form an intermediate global variable $\hat{\mbf{\lambda}}_k$ for $\mathcal{C}_t$ by
	   		\begin{equation}
	   		  \hat{\lambda}_{kj} = \eta + \frac{D}{S} \sum_{\mbf{d} \in \mathcal{C}_t} d_j \phi_{djk}.
	   		\end{equation}
	   	\STATE Update the global variable to be a weighted average of  $\hat{\mbf{\lambda}}$ and $\mbf{\lambda}^{(t - 1)}$ by 
   			\begin{equation}
   			  \mbf{\lambda}^{(t)} := (1-\rho_t) \mbf{\lambda}^{(t-1)} + \rho_t \hat{\mbf{\lambda}}.
   			\end{equation}
   	\ENDFOR
\end{algorithmic}
\end{algorithm}

A careful observation about the algorithm reveals that in fact Online-FW is a hybrid combination of FW and variational Bayes \cite{BNJ03}, where the global variables $(\mbf{\beta})$ are approximated by variational Bayes, but the local variables $(\mbf{\theta})$ are estimated by FW. Note that our adaptation of FW to posterior inference of local variables is similar in manner with the adaptation of CGS by \cite{MimnoHB12}. One important property of Online-FW is that the quality of MAP inference of $\mbf{\theta}$ is theoretically guaranteed. In contrast, posterior inference of local variables by VB or CGS does not have any guarantee. 

\subsection{Streaming-FW for learning LDA from data streams}
A disadvantage of SVI and Online-FW is that the number of training documents has to be known a priori. In practice, one may have no way to know how many documents to be processed. In those cases, the scheme proposed by \cite{Hoffman2013SVI} cannot apply. Fortunately, \cite{Broderick2013streaming} shows a simple way to help SVI work in a real online/stream environment. 

Imagine the data come sequentially in an order. Our task is to estimate a posterior distribution from this data sequence  without knowing how many instances there are. \cite{Broderick2013streaming} suggest that we should treat the posterior of the previous data as the new prior for the incomming data points. By this way, we can estimate the posterior in a real online/stream environment. When applying this scheme to some models with conjugate priors such as LDA, saving and updating the sufficient statistics of the posterior are enough.

We now discuss how to modify Online-FW to work with data streams, following the suggestion by \cite{Broderick2013streaming}. Note that the intermediate variable $\hat{\mbf{\lambda}}$ in Algorithm~\ref{alg:-online-FW} plays the role as the variational parameters of the distribution over words with respect to the current minibatch. It contains the sufficient statistics ($\sum_{\mbf{d} \in \mathcal{C}_t}  d_j \phi_{djk}$) of the posterior of the current minibatch. Following the arguments by \cite{Broderick2013streaming}, we just need to add those statistics to  the sufficient statistics of the global posterior over topics. Nonetheless, we find that such an update of the global posterior would quickly forget the role of the prior $Dir(\eta)$ over topics, which is a crucial part that helps LDA works in practice. To maintain the regularization role of this prior in streaming environments, we propose to keep $\eta$ as a part of the sufficient statistics to be used in each minibatch. Therefore, we arrive at a new algorithm (namely, \emph{Streaming-FW}) for learning LDA as described in Algorithm~\ref{alg:-stream-FW}. 

\begin{algorithm}[tp]
   \caption{\textsf{Streaming-FW} for learning LDA}
   \label{alg:-stream-FW}
\begin{algorithmic}
   \STATE {\bfseries Input: } data sequence, hyperparameter $\eta$
   \STATE {\bfseries Output:} $\mbf{\lambda}$
   \STATE Initialize $\mbf{\lambda}^{(0)}$ randomly
   \FOR{ $t = 1, ..., \infty$}
	   \STATE Sample a set $\mathcal{C}_t$ of documents. 
	   \STATE Use Algorithm~\ref{alg:-Frank-Wolfe} to do posterior inference for each document $\mbf{d} \in \mathcal{C}_t$, given the global variable $\mbf{\beta}^{(t-1)} \propto \mbf{\lambda}^{(t - 1)}$ in the last step, to get topic mixture $\mbf{\theta}_d$. Then compute
		   \begin{equation} 
		   \phi_{djk} \propto \theta_{dk}  \beta_{kj}.
		   \end{equation}
	   \STATE For each $k \in \{1, 2, ..., K\}$, compute the sufficient statistics $\hat{\mbf{\lambda}}_k$ for $\mathcal{C}_t$ by
	   		\begin{equation}
	   		  \hat{\lambda}_{kj} = \eta + \sum_{\mbf{d} \in \mathcal{C}_t}  d_j \phi_{djk}.
	   		\end{equation}
	   	\STATE Update the global variable by
   			\begin{equation}
   			  \mbf{\lambda}^{(t)} := \mbf{\lambda}^{(t-1)} +  \hat{\mbf{\lambda}}.
   			\end{equation}
   	\ENDFOR
\end{algorithmic}
\end{algorithm}

\subsection{ML-FW for learning LDA from large corpora or data streams}

\begin{algorithm}[tp]
   \caption{\textsf{ML-FW} for learning LDA}
   \label{alg:-ML-FW}
\begin{algorithmic}
   \STATE {\bfseries Input: } data sequence, parameter $\{\kappa, \tau \}$
   \STATE {\bfseries Output:} $\mbf{\beta}$
   \STATE Initialize $\mbf{\beta}^{(0)}$ randomly in $\Delta_V$
   \FOR{ $t = 1, ..., \infty$}
	   \STATE Sample a set $\mathcal{C}_t$ of documents. 
	   \STATE Use Algorithm~\ref{alg:-Frank-Wolfe} to do posterior inference for each document $\mbf{d} \in \mathcal{C}_t$, given the global variable $\mbf{\beta}^{(t-1)}$ in the last step, to get topic mixture $\mbf{\theta}_d$. 
	   \STATE For each $k \in \{1, 2, ..., K\}$, compute the intermediate topic $\hat{\mbf{\beta}}_k$ for $\mathcal{C}_t$ by
	   		\begin{equation}
	   		  \hat{\beta}_{kj} \propto \sum_{\mbf{d} \in \mathcal{C}_t}  d_j \theta_{dk}.
	   		\end{equation}
	   	\STATE Update the global variable by, where $\rho_t = (t + \tau)^{-\kappa}$,
   			\begin{equation}
   			  \mbf{\beta}^{(t)} := (1-\rho_t) \mbf{\beta}^{(t-1)} + \rho_t \hat{\mbf{\beta}}.
   			\end{equation}
   	\ENDFOR
\end{algorithmic}
\end{algorithm}

It is worth noticing that Online-FW and Streaming-FW are  hybrid algorithms which combine FW with variational Bayes for estimating the posterior of the global variables. They have to maintain variational parameters ($\mbf{\lambda}$) for the Dirichlet distribution over topics, instead of the topics themselve. Nonetheless, the combinations are not very natural since we have to compute $\mbf{\phi}$ from $\mbf{\theta}$ in order to update the model. Such a conversion might incur some information losses.

It is more natural if we can use directly $\mbf{\theta}$ in the update of the model at each minibatch. To this end, we use an idea from \cite{ThanH2012fstm}. Instead of following Bayesian approach to estimate the distribution over topics, one can consider  the topics as parameters and estimate them directly from data. \cite{ThanH2012fstm} show that  we can estimate the topics from a given  corpus $\mathcal{C}_t$ by $\mbf{\beta}_{kj} \propto \sum_{\mbf{d} \in \mathcal{C}_t}  d_j \theta_{dk}$. Combining this with the idea of online learning \cite{Bottou1998stochastic}, one can arrive at a new algorithm (namely, \emph{ML-FW}) for learning LDA as described in Algorithm~\ref{alg:-ML-FW}.

Different from Online-FW, we need not to know a priori how many documents to be processed. Hence, ML-FW can deal well with stream/online environments in a realistical way. Note that ML-FW ignores the priors over topics ($\mbf{\beta}$) and topic mixtures ($\mbf{\theta}$) when learning LDA. This means we learn topics and topic mixtures by the maximum likelihood approach. Further, the magnitude of the global parameters ($\mbf{\lambda}$) in Online-FW and Streaming-FW can arbitrarily grow  as the data come infinitely, but the topics $\mbf{\beta}$ in ML-FW are regularized to belong to the unit simplex $\Delta_V$. Such a regularization might help ML-FW avoid overfitting. Note that due to no need of computing any matrix $\mbf{\phi}$, ML-FW would be much more efficient than both Online-FW and Streaming-FW. Those properties make ML-FW very different from Online-FW and Streaming-FW.

\section{Empirical evaluation}\label{sec:evaluation}

This section is devoted to investigating the practical behaviors of FW,  and how useful it is when FW is employed to design large-scale algorithms for learning topic models. To this end, we take the following  methods, datasets, and performance measures into investigation.

\textsc{Inference methods:}
\begin{itemize}
\item[-] \emph{Frank-Wolfe} (FW).
\item[-] \emph{Variational Bayes} (VB) \cite{BNJ03}.
\item[-] \emph{Collapsed variational Bayes} (CVB0) \cite{Asuncion+2009smoothing}. 
\item[-] \emph{Collapsed Gibbs sampling} (CGS) \cite{MimnoHB12}.
\end{itemize}

CVB0 and CGS have been observing to work best by several previous studies \cite{Asuncion+2009smoothing,MimnoHB12,Foulds2013stochastic,GaoSWYZ15}. Therefore they can be considered as the state-of-the-art inference methods. It is worth observing that VB, CVB, CVB0 never return sparse solutions or sufficient statistics (encoded by $\mbf{\gamma} -\alpha$ in Algorithm~\ref{alg:-VB}--\ref{alg:-CVB0}) when doing inference for individual documents; but CGS and FW do.

\textsc{Large-scale learning methods:}
\begin{itemize}
\item[-] Our new algorithms:  \emph{Online-FW},  \emph{Streaming-FW}, \emph{ML-FW}
\item[-] \emph{Online-CGS} by \cite{MimnoHB12}
\item[-] \emph{Online-CVB0} by \cite{Foulds2013stochastic}
\item[-] \emph{Online-VB} by \cite{Hoffman2013SVI}, which is often known as SVI
\item[-] \emph{Streaming-VB} by \cite{Broderick2013streaming} with original name to be SSU
\end{itemize}

Online-CGS \cite{MimnoHB12} is a hybrid algorithm, for which CGS is used to estimate the distribution of local variables ($\mbf{z}$) in a document, and VB is used to estimate the distribution of global variables ($\mbf{\lambda}$). Online-CVB0 \cite{Foulds2013stochastic} is an online version of the batch algorithm by \cite{Asuncion+2009smoothing}, where local inference for a document is done by CVB0. Online-VB \cite{Hoffman2013SVI} and Streaming-VB \cite{Broderick2013streaming} are two stochastic algorithms for which local inference for a document is done by VB. To avoid any possible bias in our investigation, we wrote  those 6 methods by Python in a unified framework with our best efforts, and Online-VB was taken from \url{http://www.cs.princeton.edu/~blei/downloads/onlineldavb.tar}.

\textsc{Data for experiments:} The following two large corpora were used in our experiments. \textit{Pubmed} consisting of 8.2 millions of medical articles from the pubmed central; \textit{New York Times} consisting of 300K news.\footnote{
The data were retrieved from \url{http://archive.ics.uci.edu/ml/datasets/}} The vocabulary size ($V$) of each corpus is more than 110,000. For each corpus we set aside randomly 1000 documents for testing, and used the remaining for learning.

\textsc{Parameter settings:} 
\begin{itemize}
\item[-] \emph{Model parameters:} $K=100, \alpha = 1/K, \eta = 1/K$ which were frequently used in previous studies \cite{GriffithsS2004,Foulds2013stochastic,Hoffman2013SVI}.
\item[-] \emph{Inference parameters:} at most 50 iterations were allowed  for FW and VB to do inference. We terminated VB if the relative improvement of the lower bound on likelihood is not better than $10^{-4}$. 50 samples were used in CGS for which the first 25 were discarded and the remaining were used to approximate the posterior distribution. 50 iterations were used to do inference in CVB0, in which the first 25 iterations were burned in. Those number of samples/iterations are often enough to get a good inference solution, according to \cite{MimnoHB12,Foulds2013stochastic}. 
\item[-] \emph{Learning parameters:} minibatch size $S = | \mathcal{C}_t | =5000$, $\kappa = 0.9, \tau=1$. This choice of learning parameters has been found to result in competitive performance of Online-VB \cite{Hoffman2013SVI}, Online-CVB0 \cite{Foulds2013stochastic}. Therefore it was used in our investigation to avoid any possible bias. We used default values for some other parameters in Online-CVB0.
\end{itemize}

\textsc{Performance measures:} We used \textit{NPMI} and \textit{Predictive Probability} to see the performance of the learning methods. NPMI \cite{Lau2014npmi} measures the semantic quality of individual topics. From extensive experiments, \cite{Lau2014npmi} found that NPMI agrees well with human evaluation on the interpretability and coherence of topic models. Predictive probability \cite{Hoffman2013SVI} measures the predictiveness and generalization of a model  to new data. Detailed descriptions of these measures are presented in Appendix \ref{appendix--perp}.

\subsection{Sparsity and time by  inference methods}

Inference time is the focus in our first investigation with inference methods including FW, VB, CVB0, and CGS. In order to help us see how fast they are, we used ML-FW, Online-VB, Online-CVB0, and Online-CGS to learn LDA from the two datasets; and then calculated the average time per document that FW, VB, CVB0, and CGS respectively do inference. 

\begin{figure*}[tp]
\centering
  \includegraphics[width=0.6\textwidth]{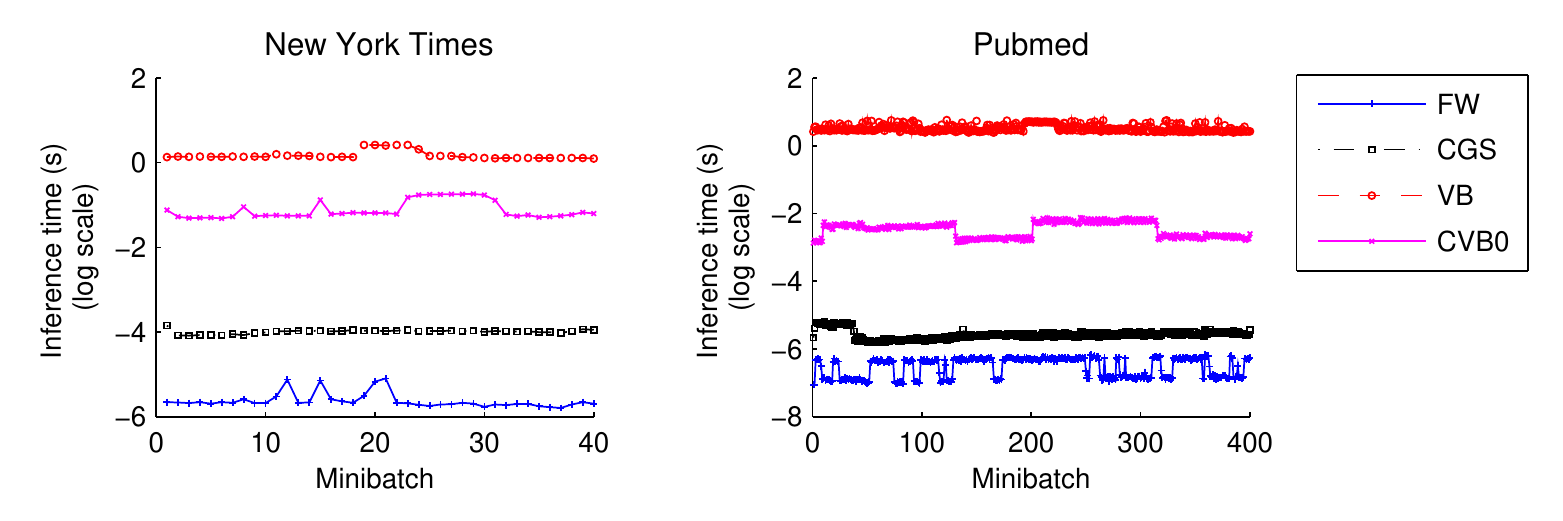} \\
  \caption{Average time to do inference for a doccument as the number of minibatches increases. Lower is faster. FW does many times faster than CGS, tens times faster than CVB0, and hundreds times faster than VB.}
  \label{fig=inference-time}
\end{figure*}

Figure \ref{fig=inference-time} shows the speed of 4 methods. We observe that FW worked fastest, followed by CGS. VB and CVB0 required significant computation time to do inference. For example, VB did approximately 390 times more slowly than FW, while CVB0 did 70 times more slowly than FW on New York Times. Such a slow inference of CVB0 and VB is due to various reasons. Remember that VB requires many evaluations of the Digamma, logarit, and exponent functions which are often expensive (see Table~\ref{table 1: theoretical comparison}). Further, VB has to check convergence when doing inference which was observed to be extremely expensive. That is why VB consumed intensive time in our experiments. 

CGS worked much faster than CVB0 and VB owing to the ability of sparse updates to the counts from samples and owing to few evaluations of Digamma/exponent functions. Although CVB0 requires no evaluation of expensive functions, it has to update all the local and global  parameters ($\mbf{\gamma, \phi, N}$) with respect to each token in the inferred document. Therefore in total the number of computations may increase very quickly if the length of documents is high. That is why CVB0 often works significantly more slowly than CGS and FW. Different from other methods, FW just requires a computation of the gradient vectors of the log likelihood and then an update of the solution. Hence FW worked very fast as depicted in Figure~\ref{fig=inference-time}.

\begin{figure*}[tp]
\centering
  \includegraphics[width=0.6\textwidth]{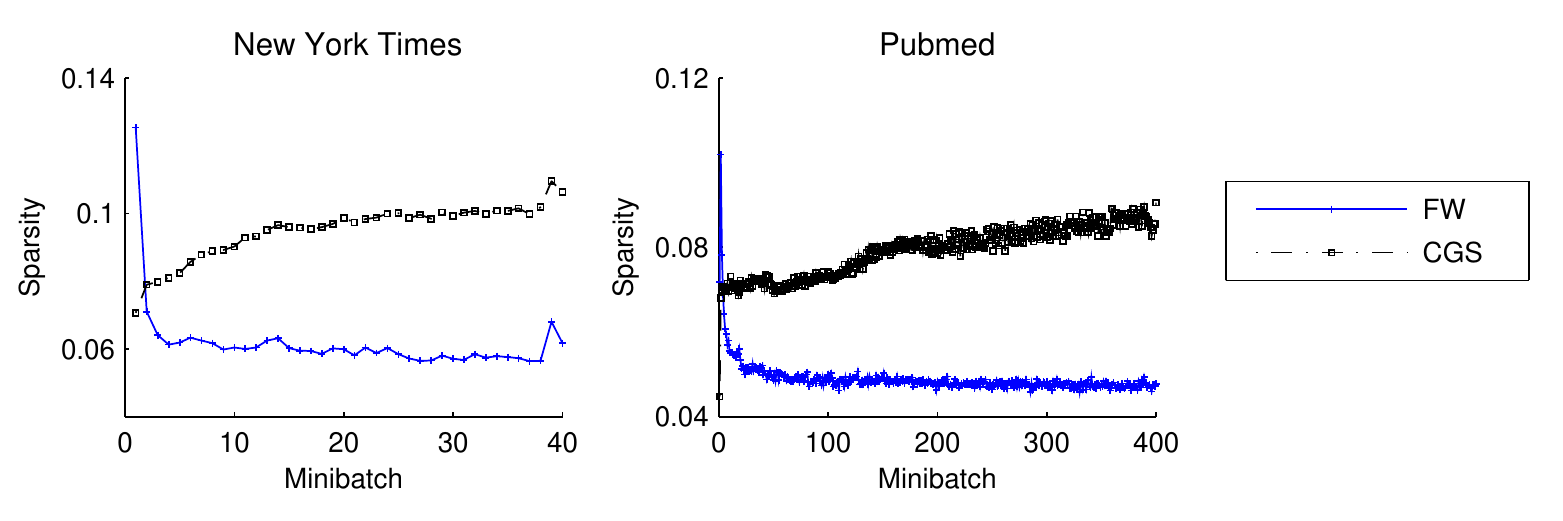} \\
  \caption{Sparsity of solutions found by FW and CGS as the number of minibatches increases. Lower is sparser.}
  \label{fig=sparsity-fw-cgs}
\end{figure*}

We next want to see \emph{how sparse are the solutions returned by the inference methods?} Sparsity refers to the number of topics appearing in a document which are inferred by an inference method. Note that a document often relates to few topics, therefore sparsity measures the fitness of inference results on real texts. It was computed as the fraction of the number of nonzero elements in $\mbf{\theta}$ or $\mbf{\gamma}$ in Algorithms \ref{alg:-VB}--\ref{alg:-Frank-Wolfe}. It is worth noting that VB, CVB, and CVB0 never return sparse solutions; but CGS does without accounting for the hyperparameter $\alpha$. 

Figure \ref{fig=sparsity-fw-cgs} shows sparsity of FW and CGS, for which we counted the number of topics in $\mbf{\theta}$ in FW and the number of topics appearing in samples of CGS. We see that both methods can find sparse solutions/statistics. It is worth noting that  on average  FW inferred 5-7 topics while CGS inferred 8-10 topics per document. A text written by human often  talks about few topics. It suggests that 8-10 topics in a document seems to be unrealistic. Furthermore, Figure~\ref{fig=sparsity-fw-cgs} tells that the solutions by CGS tends to be denser as continuing learning which is unrealistic. In contrast, on average the sparsity in FW is quite stable as continuing learning, and inference of 5-7 topics seems to better fit with common texts. From those observations, FW seems to be better than CGS in both  sparsity and fitness with real texts.

\begin{figure*}[tp]
\centering
  \includegraphics[width=0.6\textwidth]{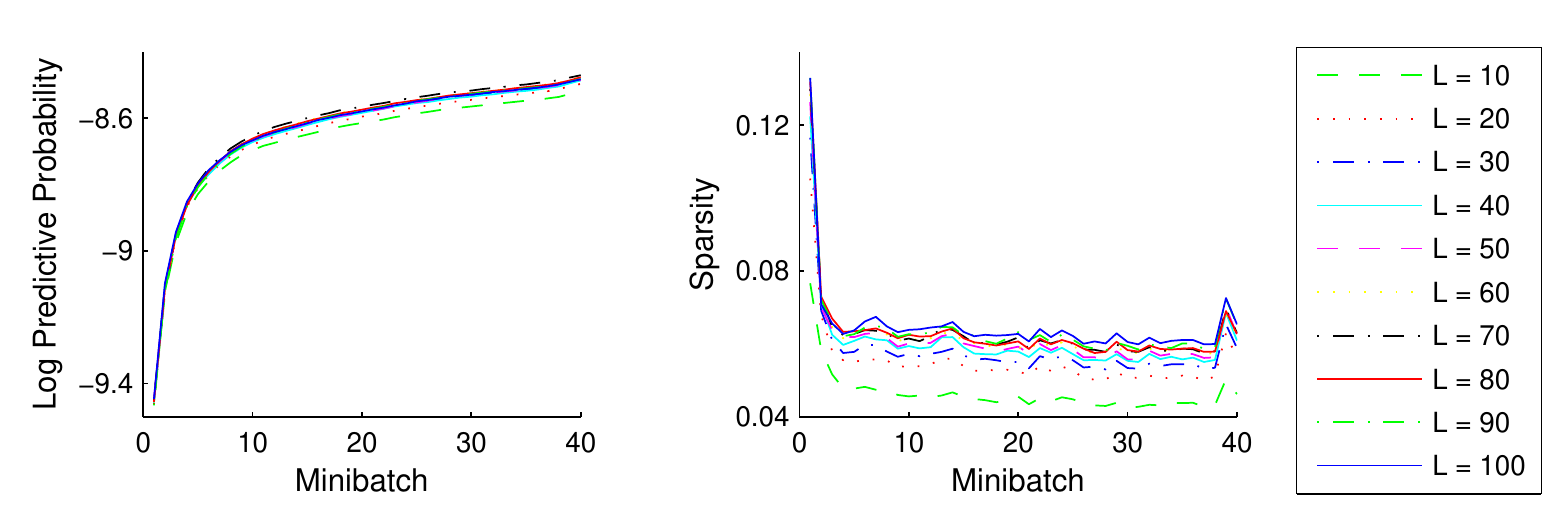} \\
  \caption{The effects of the number $L$ of iterations in FW. News York Times and ML-FW were taken in this investigation.}
  \label{fig=sensitivity-fw}
\end{figure*}

\emph{Convergence rate of FW:} We have seen that FW does inference very fast, compared with existing methods. Our next investigation is to see how fast FW converges to the optimal solution in practice. Theorem~\ref{thm-FW} ensures a linear rate of convergence for FW. Figure~\ref{fig=sensitivity-fw} tells us more about performance of FW in practice. We observe that more iterations may lead to denser solutions, but do not infer significant dense solutions. When FW is encoded in ML-FW for learning LDA, Figure~\ref{fig=sensitivity-fw} shows that allowing more iterations for FW does not always get better models. 30 iterations seem to be enough for FW to help us learn a good model, since there was no statistically significant difference in predictiveness for different settings as the  number of iterations is at least 20. Those observations suggest that FW converges very fast in practice.

\subsection{Performance of learning algorithms}

In this section, we investigate the performance of our new algorithms for learning LDA at large scales, and the benefits when employing FW to do posterior inference in topic models. We took 4 existing methods into investigation including Online-CGS, Online-CVB0, Online-VB and Streaming-VB. Following previous studies, we set $\alpha = 1/K = 0.01$ for the Dirichlet prior over $\mbf{\theta}$ to get competitive performance for those four methods. Remember that employing FW implies the use of Dirichlet prior with $\alpha=1$ in ML-FW, Online-FW, and Streaming-FW. It means that our new methods learn a different LDA model. Therefore, to make a better comparison, we also did experiments with Online-CGS, Online-CVB0, Online-VB and Streaming-VB for the case of $\alpha=1$.

\subsubsection{Predictiveness}

\begin{figure*}[tp]
\centering
  \includegraphics[width=0.8\textwidth]{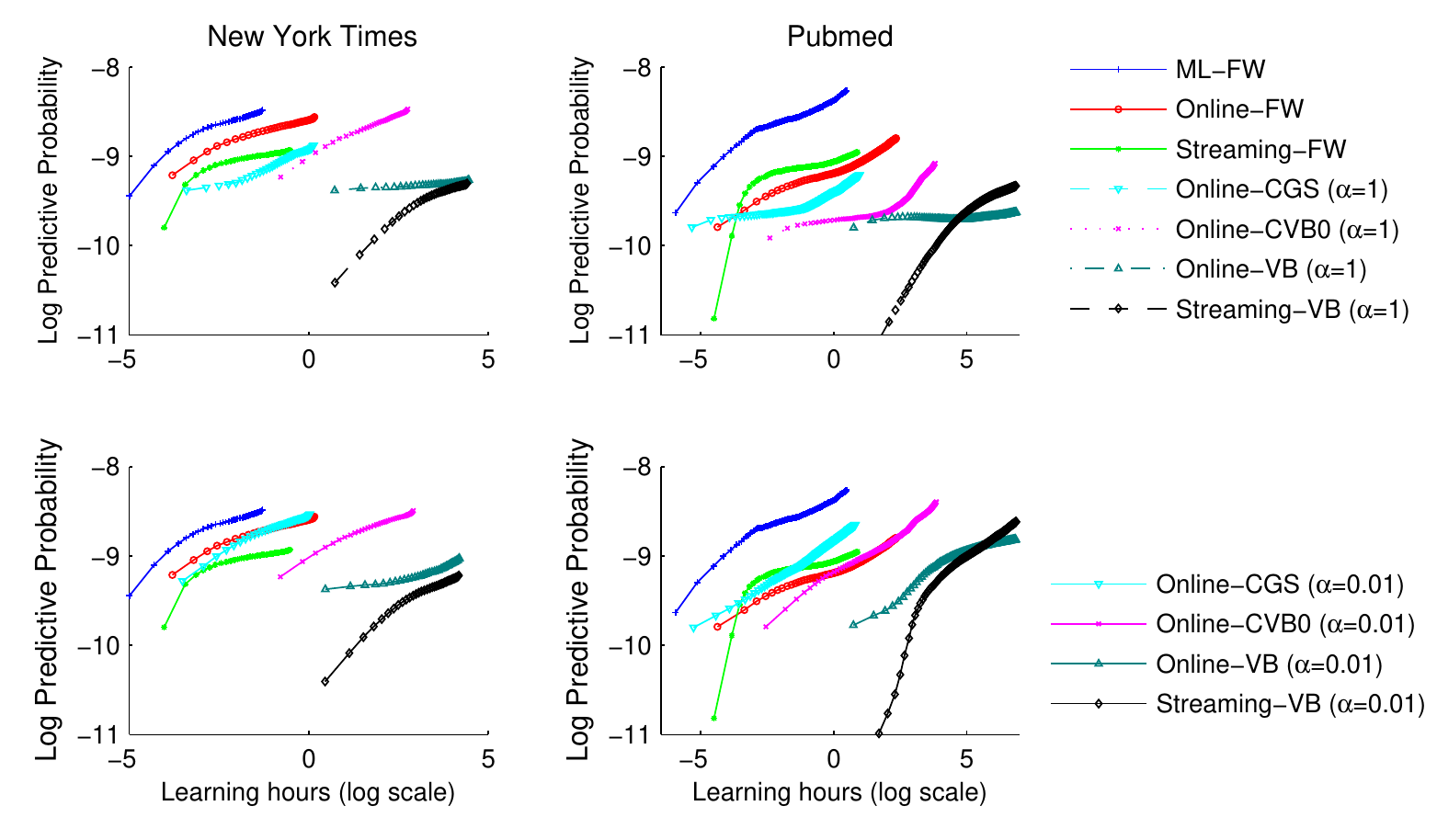} \\
  \caption{Predictiveness of models learned by different methods as spending more learning time. Higher is better. The first row shows the methods which learned LDA with hyperparameter $\alpha=1$. The second row shows the methods for learning LDA in the case of $\alpha=0.01$. It can be observed that to reach the same predictiveness level, ML-FW performs tens times faster than Online-CGS, hundred times faster than Online-CVB0, and thousands times faster than both Online-VB and Streaming-VB.}
  \label{fig=perplexity-learning-methods}
\end{figure*}

Figure \ref{fig=perplexity-learning-methods} depicts the performance of 11 learning methods as spending more time for learning. Observing the figure we see that ML-FW, Online-FW,  Streaming-FW, and Online-CGS are among the most efficient methods. They reached very quickly to a good predictiveness level. To reach to the same  level, other methods required substantially more learning time. It is worth noticing that for the same LDA model with $\alpha=1$, FW-based methods got higher predictiveness level than the others. This result suggests that FW can do inference significantly better than VB, CVB0, and CGS for the same models.

In the case of $\alpha=0.01$, Online-CGS and Online-CVB0 can reach to a very high predictiveness level, which agree well with previous studies \cite{MimnoHB12,Foulds2013stochastic,GriffithsS2004}. Online-VB and Streaming-VB can perform well, but with intensive learning time due to the expensive computation of VB. Among 11 methods for learning LDA, the following three reached to top performance: ML-FW, Online-CGS, and Online-CVB0. It is easy to observe from Figure~\ref{fig=perplexity-learning-methods} that  Online-CVB0 required significantly more time than Online-CGS and ML-FW. The reason comes from the intensive computation of CVB0 as analyzed before. Both FW and CGS consumes light computation, and hence they can help ML-FW and Online-CGS learn very fast.

It is worth noticing that ML-FW performed best among 11 methods on both New York Times and Pubmed. For a given learning time budget, ML-FW often reached to a very high predictiveness level, compared with other methods. The superior performance of ML-FW might come from the facts that the solutions ($\mbf{\theta}$)  from FW are provably good, and that the quality of solutions from FW are inherited  directly in ML-FW to update the global variables ($\mbf{\beta}$). VB, CVB0, and CGS do not have any guarantee on quality, and may require a large number of iterations/samples to get a good solution. Unlike ML-FW, Online-FW and Streaming-FW do not always perform better than other methods. The reasons might come from the indirect use of $\mbf{\theta}$ to update the global variables ($\mbf{\lambda}$). The indirect use of qualified $\mbf{\theta}$ in Online-FW and Streaming-FW might incur some losses. This could be one of the main reasons for the inferior performance of those two methods.

\subsubsection{Semantic quality}

\begin{figure*}[tp]
\centering
  \includegraphics[width=0.8\textwidth]{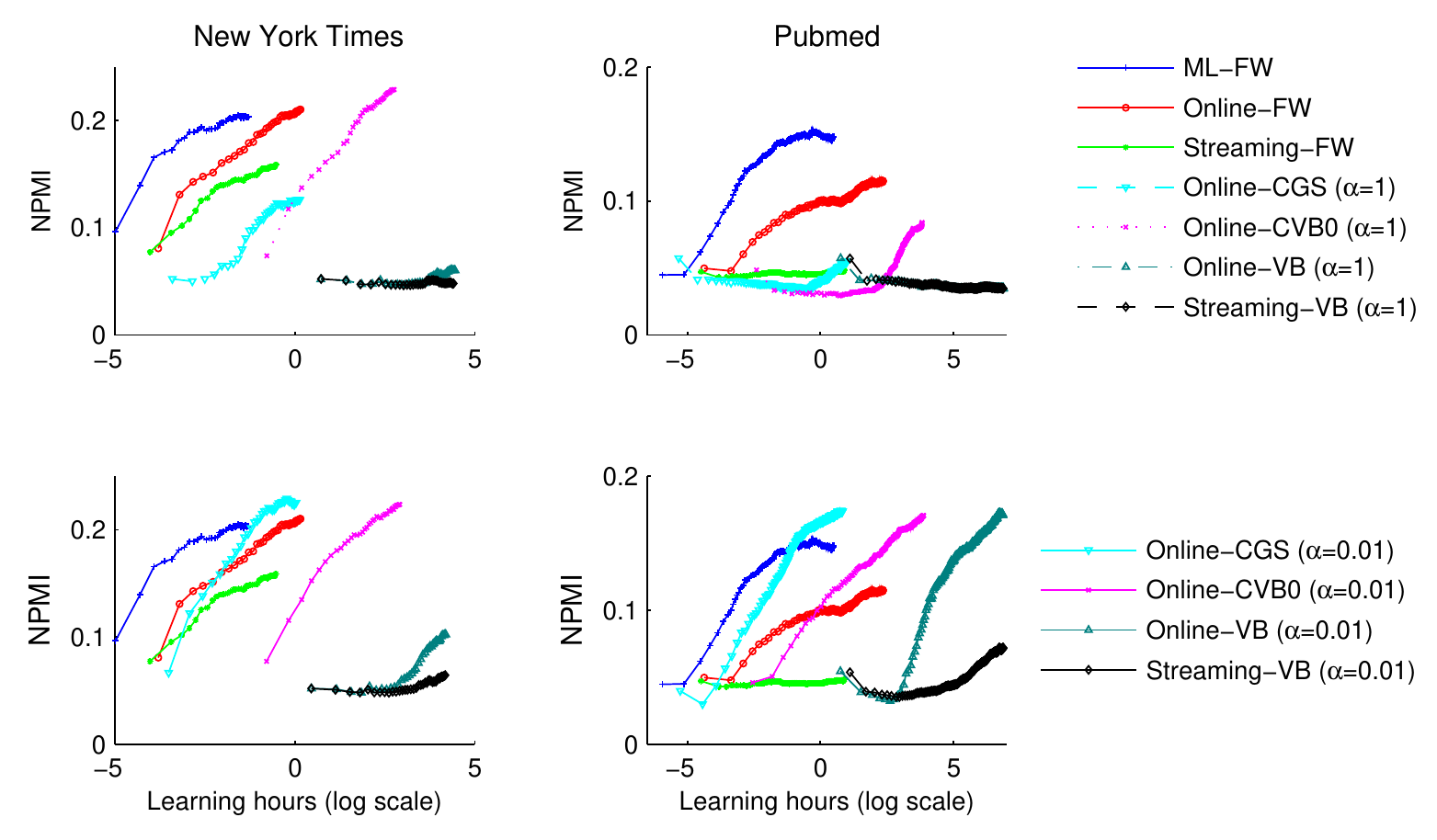} \\
  \caption{Quality of LDA learned by different methods as spending more learning time. Higher is better. The first row shows the methods which learned LDA with hyperparameter $\alpha=1$. The second row shows  the case of $\alpha=0.01$.}
  \label{fig=npmi-learning-methods}
\end{figure*}

We next want to see the semantic quality of the models learned by different methods. We used NPMI as a standard measure, because it has been observed to agree well with human evaluation on interpretability of topics. Figure~\ref{fig=npmi-learning-methods} presents the results of 11 methods.

Similar with predictiveness, FW-based methods often resulted in better models than the other methods when the same models ($\alpha=1$) are in consideration. ML-FW and Online-FW did consistently better than Streaming-FW. It seems that $\alpha=1$ is not the good condition for the traditional inference methods such as VB, CVB0, and CGS to do inference. On contrary, FW exploits well this condition to optimally infer  topic proportions ($\mbf{\theta}$). That might be why FW-based learning methods performed significantly better than the others.

Among 11 learning methods and in unrestricted settings (such as $\alpha=0.01$), Online-CVB0 seems to perform best if it is allowed enough learning time. Online-VB and Streaming-VB often work very slowly, while FW-based methods and Online-CGS can quickly learn a good LDA model. It is interesting that Online-CVB0 performed well with respect to both measures (Predictive Probability and NPMI). The reasons might come from the facts that CVB0 helps us better approximate the likelihood than VB \cite{TehNW2007collapsed,Asuncion+2009smoothing,SatoN2012CVB0}, and that the ability to exploit individual tokens can help CVB0 infer better. Our experimental results here agree well with previous studies on CVB0 and CGS \cite{Foulds2013stochastic,MimnoHB12,Asuncion+2009smoothing,GaoSWYZ15,SatoN2015SCVB0}.

\begin{figure*}[tp]
\centering
  \includegraphics[width=0.8\textwidth]{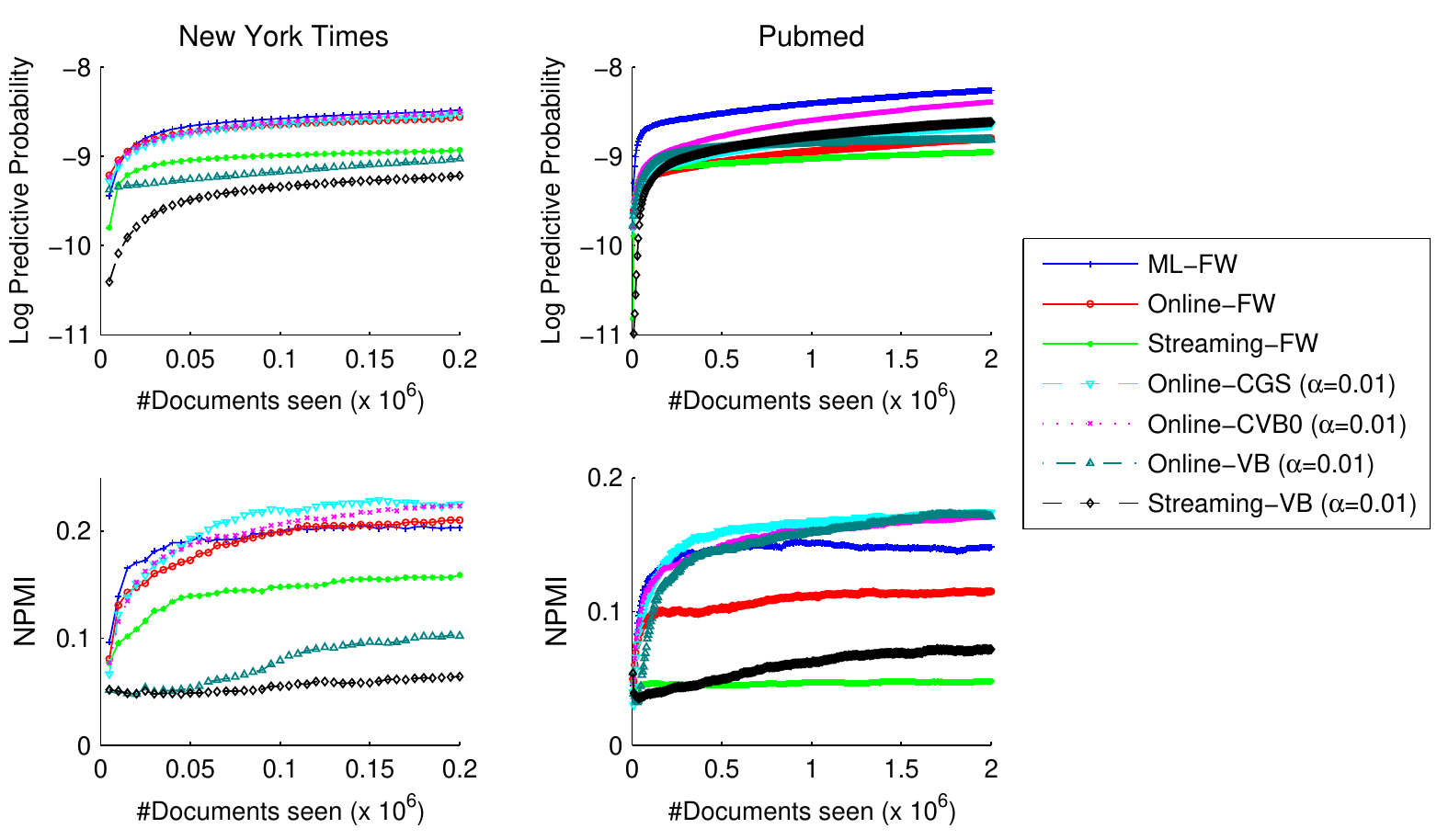} \\
  \caption{Performance of  different methods as seeing more documents. Higher is better.}
  \label{fig=perp-npmi-learning-methods}
\end{figure*}

Figure \ref{fig=perp-npmi-learning-methods} shows another perspective on performance of the large-scale learning methods. We find that ML-FW, Online-FW, Online-CGS, and Online-CVB0 can reach to a high predictiveness level just after seeing 100K documents. More texts always improve their predictiveness. In terms of semantic quality (NPMI), ML-FW were often among the top performers but neither Streaming-FW nor Online-FW. For New York Times, Online-FW could outperform the others. However, the performance of Online-FW was not very stable to reach top performance. Some information losses could incur when recovering $\mbf{\phi}$ from $\mbf{\theta}$ in Online-FW and Streaming-FW.

In summary, Figures \ref{fig=perplexity-learning-methods}--\ref{fig=perp-npmi-learning-methods} show that ML-FW and Online-FW can reach to comparable performance with state-of-the-art methods for learning LDA. ML-FW often outperforms the others in both efficiency and effectiveness (predictiveness). Those results clearly illustrate   practical benefits of FW.

\section{Conclusion}
We have investigated the use of the Frank-Wolfe algorithm (FW) \cite{Clarkson2010} to do posterior inference in topic modeling. By detailed comparisons with existing inference methods in both theoretical and practical perspectives, we elucidated many interesting benefits of FW when employed in topic modeling and in large-scale learning. FW is theoretically guaranteed on inference quality, can swiftly infer sparse solutions, and enable us to easily design efficient large-scale methods for learning topic models. Our investigation resulted in 3 novel stochastic methods for learning LDA at large scales, among which ML-FW  reaches state-of-the-art performance. ML-FW can work with big collections and text streams, and therefore provides  a new efficient tool to the public community. The code of those methods is available at \url{http://github.com/Khoat/OPE/.}


%

\appendices

\section{Predictive Probability}
\label{appendix--perp}

Predictive Probability shows the predictiveness and generalization of a model $\mathcal{M}$ on new data. We followed the procedure in \cite{Hoffman2013SVI} to compute this quantity. For each document in a testing dataset, we divided randomly into two disjoint parts $\mbf{w}_{obs}$ and $\mbf{w}_{ho}$ with a ratio of 80:20. We next did inference for $\mbf{w}_{obs}$ to get an estimate of $\mathbb{E}(\mbf{\theta}^{obs})$. Then we approximated the predictive probability as 
\[
\Pr(\mbf{w}_{ho} | \mbf{w}_{obs}, \mathcal{M}) \approx \prod_{w \in \mbf{w}_{ho} } \sum_{k=1}^{K} \mathbb{E}(\mbf{\theta}^{obs}_k) \mathbb{E}(\mbf{\beta}_{kw}),
\]
\[
\text{Log Predictive Probability} = \frac{\log \Pr(\mbf{w}_{ho} | \mbf{w}_{obs}, \mathcal{M})}  {|\mbf{w}_{ho}|},
\]
where $\mathcal{M}$ is the model to be measured. We estimated $\mathbb{E}(\mbf{\beta}_k) \propto \mbf{\lambda}_k$ for the learning methods which maintain a variational distribution ($\mbf{\lambda}$) over topics. Log Predictive Probability was averaged from 5 random splits, each was on 1000 documents.

\section{NPMI}

\emph{NPMI}  \cite{Aletras2013evaluating,Bouma2009NPMI} is the measure to help us see the coherence or semantic quality of individual topics. According to \cite{Lau2014npmi}, NPMI  agrees well with human evaluation on interpretability of topic models. For each topic $t$, we take the set $\{w_1, w_2, ..., w_n\}$ of top $n$ terms with highest probabilities. We then computed

\[
NPMI(t) = \frac{2}{n(n-1)} \sum_{j=2}^{n} \sum_{i=1}^{j-1} \frac{\log \frac{P(w_j, w_i)}{P(w_j) P(w_i)}}{- \log P(w_j, w_i)},
\]
where $P(w_i, w_j)$ is the probability that  terms $w_i$ and $w_j$ appear together in a document. We estimated those probabilities from the training data. In our experiments, we chose top $n=10$ terms for each topic.

Overall, NPMI of a model with $K$ topics is averaged as:
\[
NPMI = \frac{1}{K} \sum_{t=1}^{K} NPMI(t).
\]

\ifCLASSOPTIONcompsoc
  \section*{Acknowledgments}
\else
  \section*{Acknowledgment}
\fi

This work was partially supported by Vietnam National Foundation for Science and Technology Development (NAFOSTED Project No. 102.05-2014.28), and by AOARD (U.S. Air Force) and ITC-PAC (U.S. Army)  under agreement number FA2386-15-1-4011.

\ifCLASSOPTIONcaptionsoff
  \newpage
\fi



\bibliographystyle{IEEEtran}
\bibliography{../topic-models-all,../other-all}
%

%

\begin{IEEEbiography}[{\includegraphics[width=1in,height=1.25in,clip,keepaspectratio]{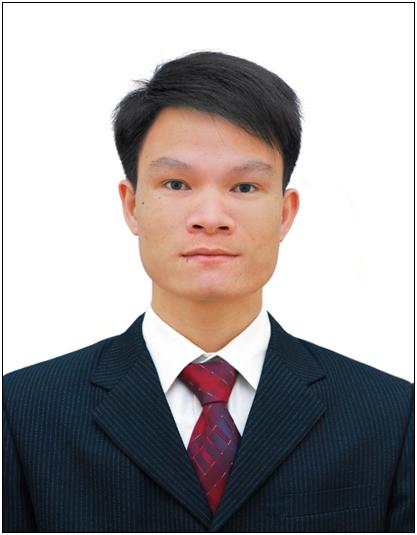}}]{Khoat Than}
 received B.A in Applied Mathematics and Informatics (2004) from Vietnam National University,  M.A in Computer Science (2009) from Hanoi University of Science and  Technology, and Ph.D in Computer Science (2013) from Japan Advanced Institute of Science and Technology. His research interests include topic modeling, dimension reduction, manifold learning, large-scale modeling, graphical models.
\end{IEEEbiography}

\begin{IEEEbiography}[{\includegraphics[width=1in,height=1.25in,clip,keepaspectratio]{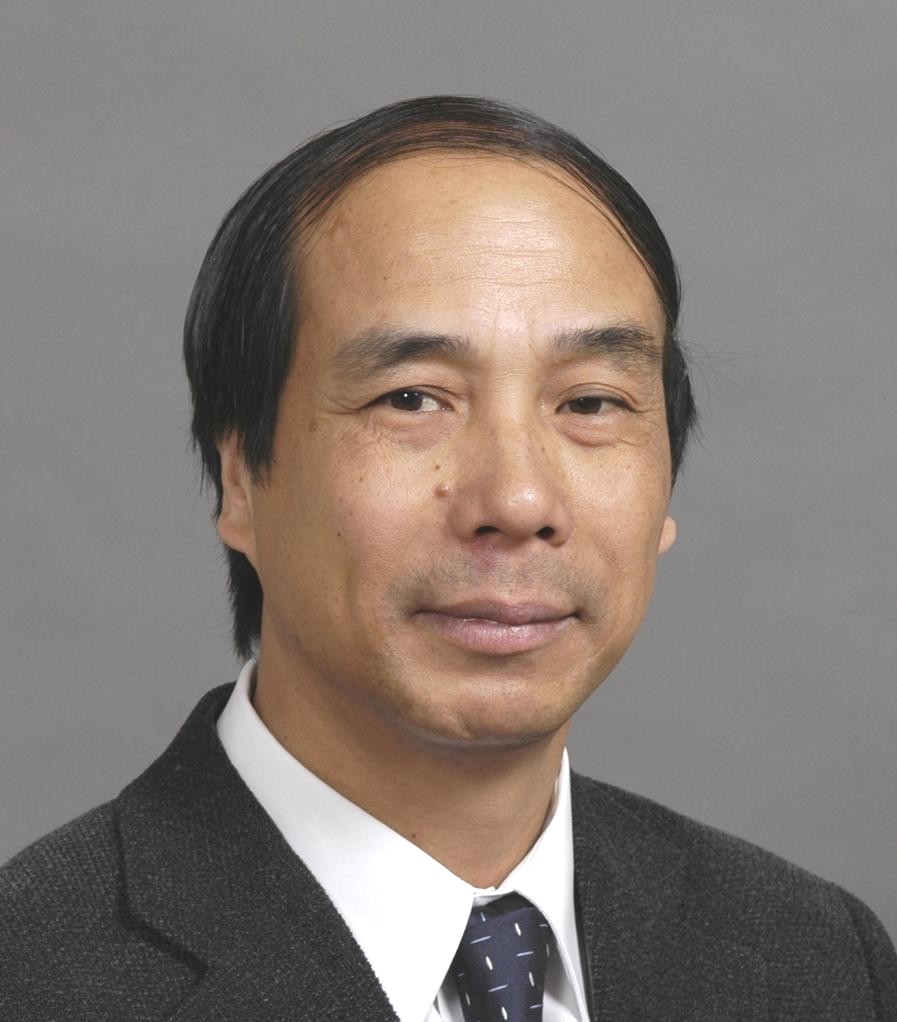}}]{Tu Bao Ho}   is currently is a professor of School of Knowledge Science, Japan Advanced Institute of Science and Technology. He received a BT in applied mathematics from Hanoi University of Science and  Technology (1978), MS and PhD in computer science from Pierre and Marie Curie University, Paris (1984, 1987). His research interests include knowledge-based systems, machine learning, knowledge discovery and data mining.
\end{IEEEbiography}






\end{document}